\documentclass[conference]{IEEEtran}
\IEEEoverridecommandlockouts
\usepackage{cite}
\usepackage{amsmath,amssymb,amsfonts}
\usepackage{algorithmic}
\usepackage{graphicx}
\usepackage{textcomp}
\usepackage{xcolor}
\usepackage{multirow}
\usepackage{geometry}
\def\BibTeX{{\rm B\kern-.05em{\sc i\kern-.025em b}\kern-.08em
    T\kern-.1667em\lower.7ex\hbox{E}\kern-.125emX}}

\newgeometry{left=54pt, right=54pt, top=72pt, bottom=54pt}
\begin{document}

\newgeometry{left=54pt, right=54pt, top=72pt, bottom=54pt}
\title{\LARGE \bf
Deep RL-based Autonomous Navigation of Micro Aerial Vehicles (MAVs) in a complex GPS-denied Indoor Environment
}


\author{Amit Kumar Singh$^{1}$, Prasanth Kumar Duba$^{2}$ and P. Rajalakshmi$^{3}$
\thanks{$^{1}$Dept. of Artificial Intelligence,
TiHAN, IIT Hyderabad, India.
        {\tt\small ai22mtech12010@iith.ac.in}}%
\thanks{$^{2}$Dept. of Electrical Engineering, TiHAN, IIT Hyderabad, India.
        {\tt\small ee20resch11008@iith.ac.in}}%
\thanks{$^{3}$Dept. of Electrical Engineering, TiHAN, IIT Hyderabad, India.
        {\tt\small raji@ee.iith.ac.in}}%
}

\maketitle
\begin{abstract}
The Autonomy of Unmanned Aerial Vehicles (UAVs) in indoor environments poses significant challenges due to the lack of reliable GPS signals in enclosed spaces such as warehouses, factories, and indoor facilities. Micro Aerial Vehicles (MAVs) are preferred for navigating in these complex, GPS-denied scenarios because of their agility, low power consumption, and limited computational capabilities. In this paper, we propose a Reinforcement Learning based Deep-Proximal Policy Optimization (D-PPO) algorithm to enhance realtime navigation through improving the computation efficiency. The end-to-end network is trained in 3D realistic meta-environments created using the Unreal Engine. With these trained meta-weights, the MAV system underwent extensive experimental trials in real-world indoor environments. The results indicate that the proposed method reduces computational latency by 91\% during training period without significant degradation in performance. The algorithm was tested on a DJI Tello drone, yielding similar results.
\end{abstract}


\section{Introduction}
\begin{figure}[t]
\centerline{\includegraphics[width=1\linewidth, height=0.2\textheight]{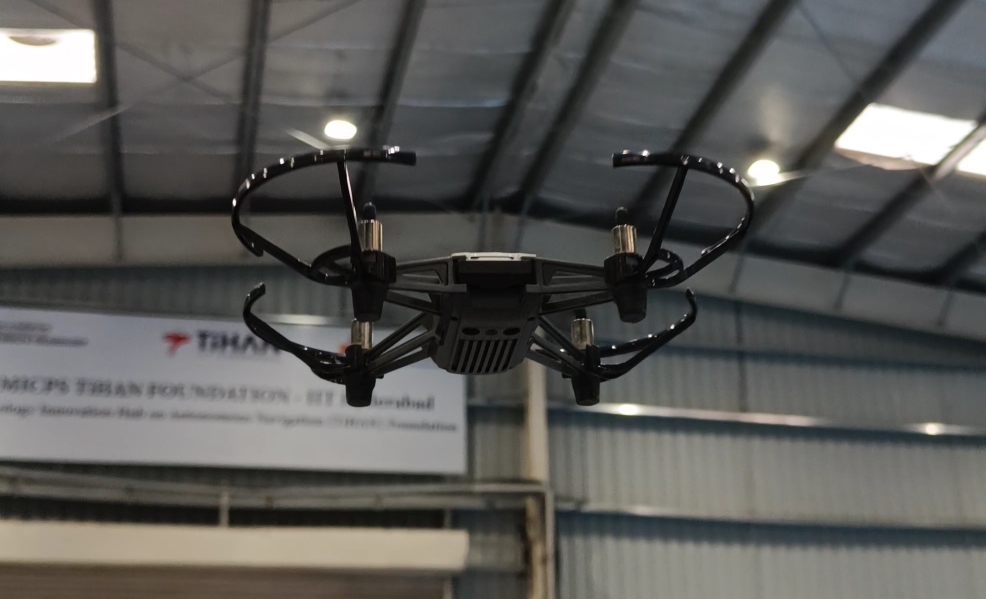}}
\caption{Autonomous Navigation of Micro Aerial Vehicle (MAV) inside the TiHAN-Testbed using a Deep-PPO based reinforcement learning algorithm.}
\label{fig:tello}
\end{figure}

Unmanned Aerial Vehicles (UAVs) have revolutionized diverse sectors, including surveillance, agriculture, disaster response, and infrastructure inspection. While outdoor UAV navigation has significantly advanced due to the widespread use of Global Positioning System (GPS) technology, developing robust and efficient navigation systems for UAVs in GPS-denied indoor environments is essential for further expanding their applications and capabilities. One of the primary hurdles in indoor UAV navigation is accurate localization within the environment. Unlike outdoor environments where GPS signals provide precise location information, indoor spaces require alternative methods for localization.Our previous work emphasizes the use of stereo vision for the localization of Micro Aerial Vehicles (MAVs) in indoor environments \cite{r6}. Additionally, navigating through indoor environments necessitates the ability to detect and avoid obstacles in real-time to ensure safe and efficient flight paths.
In this research, we propose a novel approach for GPS-denied indoor MAV navigation using deep reinforcement learning techniques. This method enables MAVs to autonomously navigate using only monocular RGB image captured by an onboard camera that is converted to depth image using Depth Anything Model (DAM). By processing the depth image input through a Convolutional Neural Network (CNN) \cite{r1} , the MAV can extract valuable features from the environment, including obstacle, free space and spatial information, to make navigation decisions in real-time. The MAV navigation in real-word indoor environment is shown in Fig.\ref{fig:tello}. The core of our approach lies in the integration of deep reinforcement learning algorithms \cite{r2} with state-of-the-art computer vision techniques, allowing the vehicle to learn optimal navigation policies directly from raw sensory inputs. By training the MAV in simulated environments, we aim to develop a navigation system capable of robust and adaptive performance across various indoor scenarios.

\section{Methodolgy}
\subsection{TiHAN-IITH Testbed}
Technology Innovation Hub on Autonomous Navigation (TiHAN) at Indian Institute if Technology Hyderabad (IITH) is sanctioned by DST-India. It is a first-of-its-kind integrated testbed which has state of the art facilities such as test tracks, hangar, ground control station , simulation tools (SIL, MIL, HIL, VIL), rainfall simulators, drone runways $\&$ landing areas\cite{r3}.

\subsection{Research Design}
This study employs a simulation-based approach using AirSim, an open-source quadrotor simulator integrated with Unreal Engine, to create realistic environments. The research utilizes state-of-the-art deep reinforcement learning techniques, specifically the Proximal Policy Optimization (PPO) algorithm, to train a convolutional neural network (CNN) for autonomous UAV navigation and collision avoidance tasks. Unlike previous methods that rely on stereo depth images\cite{r9} for their accuracy, This approach converts monocular RGB images from the drone into monocular depth images using Depth Anything Network\cite{r7}. The key insight is that while precise depth measurements are beneficial, relative depth information is sufficient for effective drone navigation.


\begin{figure}[t]
\centerline{\includegraphics[width=1\linewidth, height=0.2\textheight ]{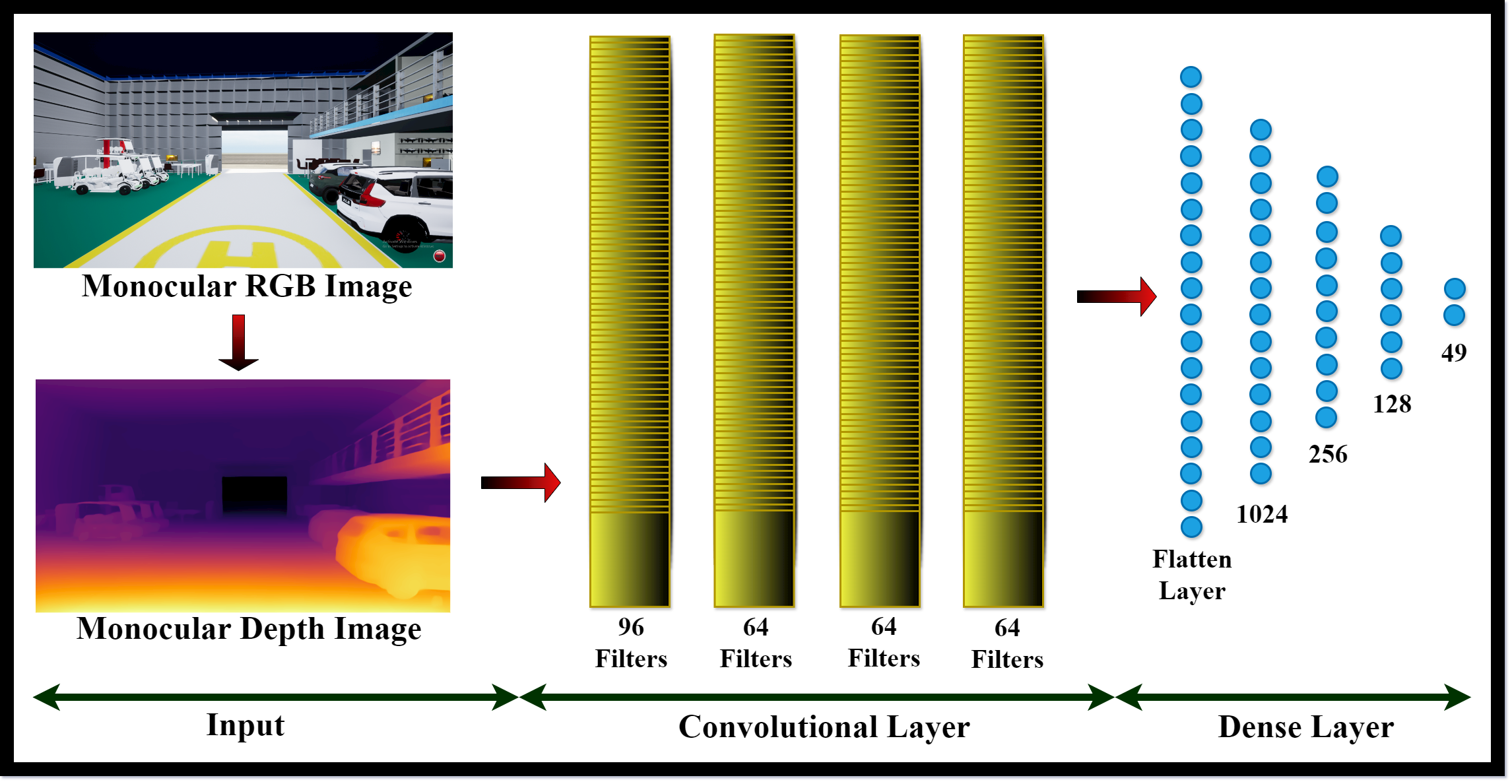}}
\caption{Proposed system architecture:  Conversion of monocular RGB image to depth image followed by CNN training with Deep-PPO algorithm.}
\label{fig:architecture}
\end{figure}

\subsection{Network Architecture and Action Spaces}
Our network processes 128×128×1 monocular depth image through a series of convolutional and max-pooling layers: starting with 96 filters (7×7), then 64 filters (5×5), and two additional layers with 64 filters each. The flattened output flows through dense layers of 1024, 256, and 128 neurons with ReLu as activation function. (Fig. \ref{fig:architecture}). The final layer, comprising 49 nodes with softmax activation, defines a 7x7 action grid for drone navigation. Each grid cell represents a potential move, enabling the drone to make informed decisions based on its training. This setup optimizes collision free navigation in the environment. The approach integrates deep learning\cite{r8} to ensure efficient and safe UAV navigation in dynamic environments, emphasizing adaptability and effective path planning capabilities.


\subsection{Training Process}
The task is to find the policy with the maximum expected Return with input as monocular depth image and output being the 7X7 Action Space. The training process using the Deep-Proximal Policy Optimization (D-PPO) algorithm\cite{r5} starts with initializing the policy (actor) and value (critic) network parameters. 
\begin{equation}
\pi^*_\theta = \arg\max_{\pi} \mathbb{E}\left[\sum_{t=0}^{\infty} \gamma^t \cdot R(t)\right]
\end{equation}

where R(t) is the Reward Function, $\gamma$ is the discount factor, $\pi$ is the policy and $\theta$ is the paramaters of the Neural Network.
The agent interacts with the environment by selecting actions according to the current policy, receiving rewards, and observing the next states. These interactions generate trajectories comprising states, actions, rewards, and next states. Using these trajectories, the advantage function is computed to evaluate the effectiveness of actions taken compared to expected actions. The policy is then updated using gradient descent, constrained to ensure that updates do not deviate excessively from the previous policy, balancing exploration and exploitation. This process iterates, continually refining the policy to improve the agent's performance in achieving its objectives while maintaining stability in learning.


\subsection{Reward Function}
In the context of reinforcement learning for UAV navigation and collision avoidance tasks, the reward function serves as a crucial component in shaping the behavior of the learning agent, guiding it towards desirable outcomes while discouraging undesirable actions. The reward function is designed to provide immediate feedback to the agent at each time step, influencing its decision-making process and learning trajectory. The reward function to navigate without any collision is defined as:
\begin{equation}
\text{R(t)} = 
\begin{cases} 
-10, & \text{On Collision} \\
\dfrac{100}{d}, & \text{Otherwise}
\end{cases}
\end{equation}

where d is the distance from the RGB image center to the center of free space in the depth image. The free space is determined by thresholding the depth image with respect to far objects and finding their center. This approach encourages safe, efficient navigation while minimizing collision risks, guiding the agent towards desirable actions and trajectories.

\section{Results and Discussion}
\subsection{Simulation Study}
\begin{figure}[h]
\centerline{\includegraphics[width=1\linewidth, , height=0.15\textheight]{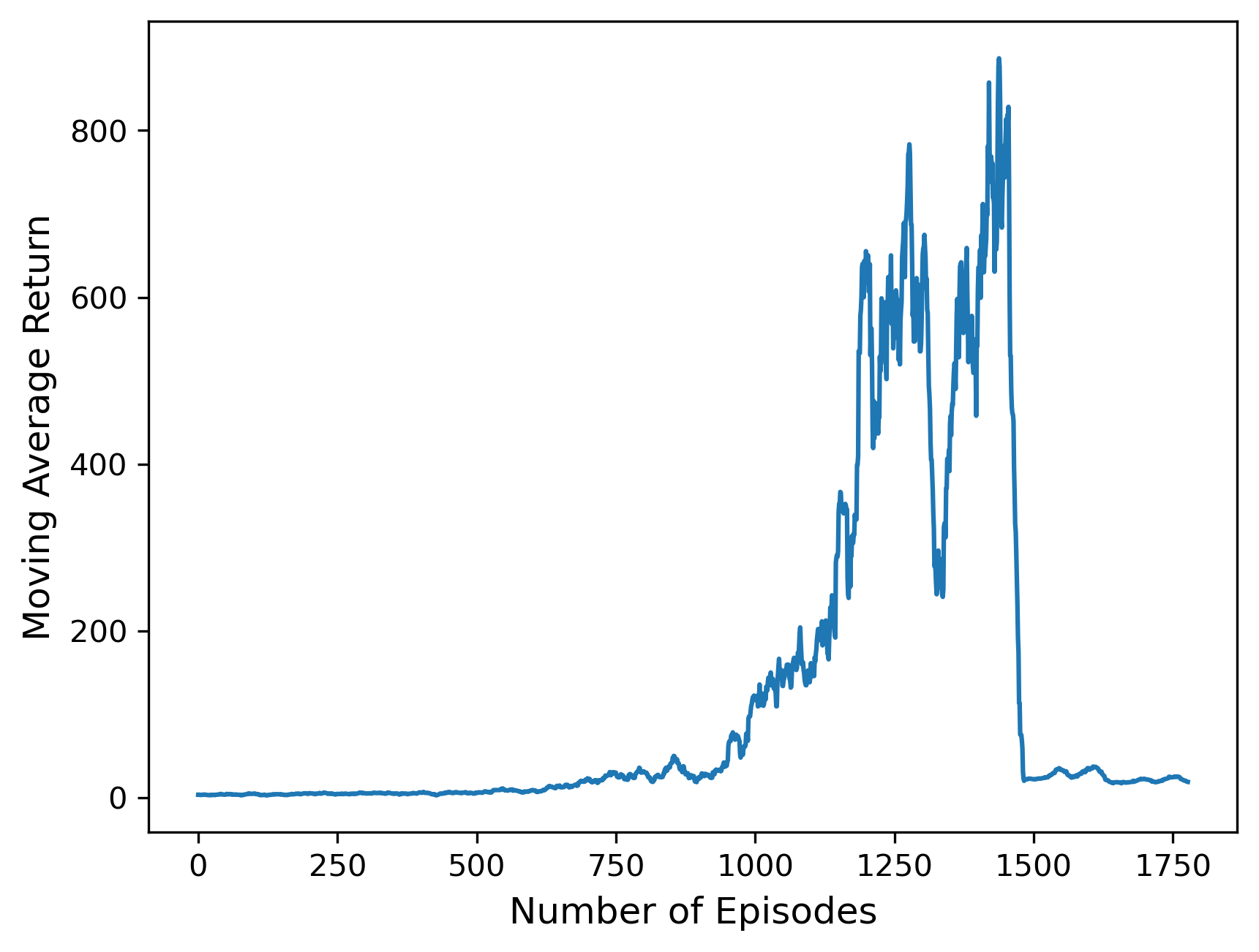}}
\caption{Plot of moving average value variations across episodes in ten simulations.}
\label{fig:return_plot}
\end{figure}

\begin{figure}[h]
\centerline{\includegraphics[width=1\linewidth, height=0.15\textheight]{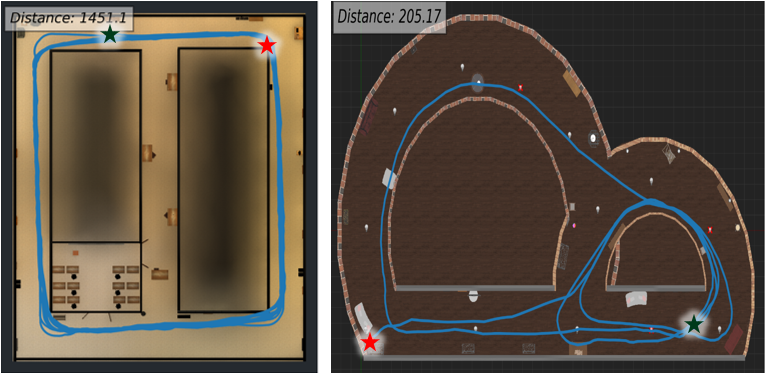}}
\caption{A 3D realistic meta simulated arena. Left : Navigation in Vanleer environment, Right : Navigation in Cloud environment.}
\label{fig:simulation}
\end{figure}

 The simulation analysis carried out for the ten experiments in a 3D realistic meta environment created by \cite{r4} as shown in Fig.\ref{fig:simulation}. The maximum and average mean safe flight travelled by the aerial vehicle in the particular environment is 1451 m and 1231 m respectively. The green star shows the starting point, while the blue line shows the flight path and collision is represented by the red star within the environment. The simulation results were comparable to the previous methods\cite{r4},\cite{r10} where they get the mean safe flight of 1245.7 m in the same environment. The system has trained with monocular depth images using Deep-PPO algorithm and the performance in terms of  variation in moving average return with episodes as shown in Fig.\ref{fig:return_plot}. The system performance is better with the maximum average return of $\simeq 800$ at $\simeq 1400$ episode. 
\newpage
The same weights are used for the experiments carried out in simulation as well in real-world scenarios.


Additionlly, the training time has been reduced drastically with the proposed method when compared to monocular RGB image. The system took 68 Hrs to get desired results, with monocular RGB images while the same system took 6 Hrs when the input is monocular depth image as tabulated in TABLE \ref{tab:training_time}.

\begin{table}[ht]  
  \centering
  \caption{Deep-PPO Algorithm: Training Time \& Efficiency}
  \begin{center}
  \label{tab:training_time}
  \begin{tabular}{c c c c}
    \hline
    \textbf{Methods} & \textbf{RGB Image} &  \textbf{Depth Image} & \textbf{Efficiency}\\
    \hline
    D-PPO (Ours) & 68 Hrs & 6 Hrs & 91\% \\
    \hline
  \end{tabular}
  \end{center}
\end{table}


  


\subsection{Real World Experimental Analysis}
The experiments conducted on DJI Tello drone inside TiHAN-Testbed with a dimensions of 30X30X8 m$^3$. The hardware platform consists of RGB camera and Wi-Fi which is controlled by a Ground Control Station (GCS). The pre-processing includes the generation of depth images using Depth Anything Model (DAM) for the captured RGB images is performed on the GCS. Based on the depth image, the trained neural network using Deep-PPO based reinforcement learning algorithm, provides the navigational commands such as forward, clockwise rotation, and counterclockwise instructions to the hardware platform. 
\begin{table}[ht]  
  \centering
  \caption{Real Time Anyalsis on Mean Safe Flight (MSF)}
  \label{tab:experimental_result}
  \begin{center}
    \begin{tabular}{c c c}
    \hline
    \textbf{Arena} & \textbf{Train Type} &  \textbf{Mean Safe Flight (m)} \\
    \hline
    Hallway & E2E (NavREN-RL) & 16.1\\
    
    Hallway & Last2 (Pedra) & 15.6   \\
    
    TiHAN-Testbed & D-PPO (Ours) & 58   \\
    \hline
  \end{tabular}
  \end{center}
  
\end{table}

The real-time experiments performed successfully in indoor environment with various obstacle densities with a iteration of ten times. These experiments demonstrated that our proposed method (Deep-PPO) consistently outperformed in real-time. The comparative analysis of mean safe flight metric as shown  in the Table \ref{tab:experimental_result}. The results shows that, our method achieved a significant advancement with the MSF of 58 m when compared to existing approaches NavREN-RL and Pedra with MSF's of 16.1 m and 15.6 m respectively.

\section{Conclusion}
The study demonstrates the effectiveness of Deep-PPO-based reinforcement learning in enabling autonomous navigation and collision avoidance for MAVs, specifically using reward functions. The integration of monocular depth images with CNN facilitated robust and adaptive navigation for the Tello drone in dynamic indoor environments. The simulation results closely matched the experimental results obtained in real-world scenarios. The proposed method excelled in real-time performance, effectively bridging the gap between simulation and practical application by reducing training time. This showcases the method's efficiency and practical feasibility in real-world scenarios. Future work will focus on advancing the development of MAV autonomy with Deep RL-based path planning algorithms to navigate in a dense object areas like forest, crowded air space, warehouses etc.

\section*{Acknowledgment}
This work was support and funded by the DST-NMICPS through the Technology Innovation Hub on Autonomous Navigation (TiHAN) at the Indian Institute of Technology Hyderabad (IIT Hyderabad).


\begin{thebibliography}{00}

\bibitem{r1} Loquercio, Antonio, Elia Kaufmann, René Ranftl, Matthias Müller, Vladlen Koltun, and Davide Scaramuzza. "Learning high-speed flight in the wild." Science Robotics 6, no. 59 (2021): eabg5810.
\bibitem{r2}Song, Yunlong, Mats Steinweg, Elia Kaufmann, and Davide Scaramuzza. "Autonomous drone racing with deep reinforcement learning." In 2021 IEEE/RSJ International Conference on Intelligent Robots and Systems (IROS), pp. 1205-1212. IEEE, 2021.
\bibitem{r3}Pawar, D.S., Singh, A., Pachamuthu, R. (2023). Connected Autonomous Vehicles (CAV) Testbed at IIT Hyderabad. In: Rastogi, R., Bharath, G., Singh, D. (eds) Recent Trends in Transportation Infrastructure, Volume 1. TIPCE 2022. Lecture Notes in Civil Engineering, vol 354. Springer, Singapore. 
\bibitem{r4}Anwar, Aqeel, and Arijit Raychowdhury. "Autonomous navigation via deep reinforcement learning for resource constraint edge nodes using transfer learning." IEEE Access 8, 2020 26549-26560.
\bibitem{r5}Schulman, John, Filip Wolski, Prafulla Dhariwal, Alec Radford, and Oleg Klimov. "Proximal policy optimization algorithms." arXiv preprint arXiv:1707.06347 (2017).
\bibitem{r6}Duba, Prasanth Kumar, Amit Kumar Singh, Sumit Sarkar, Syam Narayanan, and P. Rajalakshmi. "Stereo Vision-based Localization for Micro Aerial Vehicle (MAV) Autonomy in a GPS-denied Indoor Environments." Authorea Preprints (2024).
\bibitem{r7}Yang, Lihe, Bingyi Kang, Zilong Huang, Xiaogang Xu, Jiashi Feng, and Hengshuang Zhao. "Depth anything: Unleashing the power of large-scale unlabeled data." In Proceedings of the IEEE/CVF Conference on Computer Vision and Pattern Recognition, pp. 10371-10381. 2024. 
\bibitem{r8}AlMahamid, Fadi, and Katarina Grolinger. "Autonomous unmanned aerial vehicle navigation using reinforcement learning: A systematic review." Engineering Applications of Artificial Intelligence 115 (2022): 105321.
\bibitem{r9}Prasanth Kumar Duba, Naga Praveen Babu Mannam, and P. Rajalakshmi. "Stereo vision based object detection for autonomous navigation in space environments." Acta Astronautica 218 (2024): 326-329.
\bibitem{r10}Anwar, Malik Aqeel, and Arijit Raychowdhury. "NavREn-Rl: Learning to fly in real environment via end-to-end deep reinforcement learning using monocular images." In 2018 25th International Conference on Mechatronics and Machine Vision in Practice (M2VIP), pp. 1-6. IEEE, 2018.


\vspace{12pt}
\end{thebibliography}
\end{document}